\documentclass[10pt,twocolumn,letterpaper]{article}

\usepackage{cvpr}
\usepackage{times}
\usepackage{epsfig}
\usepackage{graphicx}
\usepackage{amsmath}
\usepackage{amssymb}
\usepackage[ruled,vlined,linesnumbered]{algorithm2e}
\usepackage[noend]{algpseudocode}
\usepackage{enumerate}
\usepackage{comment}

\usepackage{xspace}

\newcommand{\Fig}{Fig.\xspace}

\newcommand{\Tab}{Tab.\xspace}
\newcommand{\Alg}{Alg.\xspace}

\newcommand{\htx}{\hat{x}}
\newcommand{\hty}{\hat{y}}
\newcommand{\calI}{\mathcal{I}}
\newcommand{\calU}{\mathcal{U}}
\newcommand{\calL}{\mathcal{L}}

\usepackage[pagebackref=true,breaklinks=true,letterpaper=true,colorlinks,bookmarks=false]{hyperref}

\cvprfinalcopy % *** Uncomment this line for the final submission

\def\cvprPaperID{0366} % *** Enter the CVPR Paper ID here

\ifcvprfinal\fi
\begin{document}

%%%%%%%%% TITLE
\title{Generalized Intersection over Union: A Metric and A Loss for Bounding Box Regression}

\author{Hamid Rezatofighi$^{1,2}$\quad Nathan Tsoi$^1$\quad JunYoung Gwak$^1$\quad Amir Sadeghian$^{1,3}$\quad \\Ian Reid$^2$\quad Silvio Savarese$^1$\\\\
	$^1$Computer Science Department, Stanford University, United states\\
	$^2$School of Computer Science, The University of Adelaide, Australia\\
	$^3$Aibee Inc, USA\\
{\tt\small hamidrt@stanford.edu}}

\maketitle
%\thispagestyle{empty}
%%%%%%%%%%%%%%%%%%%%%%%%%%%%%%%%%%%%%%%%%%%%%%%%%%%%%%%%%%%%%%%%%%%%%%
\begin{abstract}

Intersection over Union (IoU) is the most popular evaluation metric used in the object detection benchmarks. However, there is a gap between optimizing the commonly used distance losses for regressing the parameters of a bounding box and maximizing this metric value. The optimal objective for a metric is the metric itself. In the case of axis-aligned 2D bounding boxes, it can be shown that $IoU$ can be directly used as a regression loss. However, $IoU$ has a plateau making it infeasible to optimize in the case of non-overlapping bounding boxes. In this paper, we address the weaknesses of $IoU$ by introducing a generalized version as both a new loss and a new metric. By incorporating this generalized $IoU$ ($GIoU$) as a loss into the state-of-the art object detection frameworks, we show a consistent improvement on their performance using both the standard, $IoU$ based, and new, $GIoU$ based, performance measures on popular object detection benchmarks such as PASCAL VOC and MS COCO.

\end{abstract}
% !TEX root = main.tex
\section{Introduction}

Bounding box regression is one of the most fundamental components in many 2D/3D computer vision tasks. Tasks such as object localization, multiple object detection, object tracking and instance level segmentation rely on accurate bounding box regression. The dominant trend for improving performance of applications utilizing deep neural networks is to propose either a better architecture backbone~\cite{liu2018path, retina} or a better strategy to extract reliable local features~\cite{he2017mask}. However, one opportunity for improvement that is widely ignored is the replacement of the surrogate regression losses such as $\ell_1$ and $\ell_2$-norms, with a metric loss calculated based on Intersection over Union ($IoU$).

\begin{figure}[t]
\begin{minipage}[b]{.99\linewidth}
	\begin{center}
		\includegraphics[width=.8\linewidth,trim={1.5cm 7.9cm 1.5cm 11.2cm},clip]{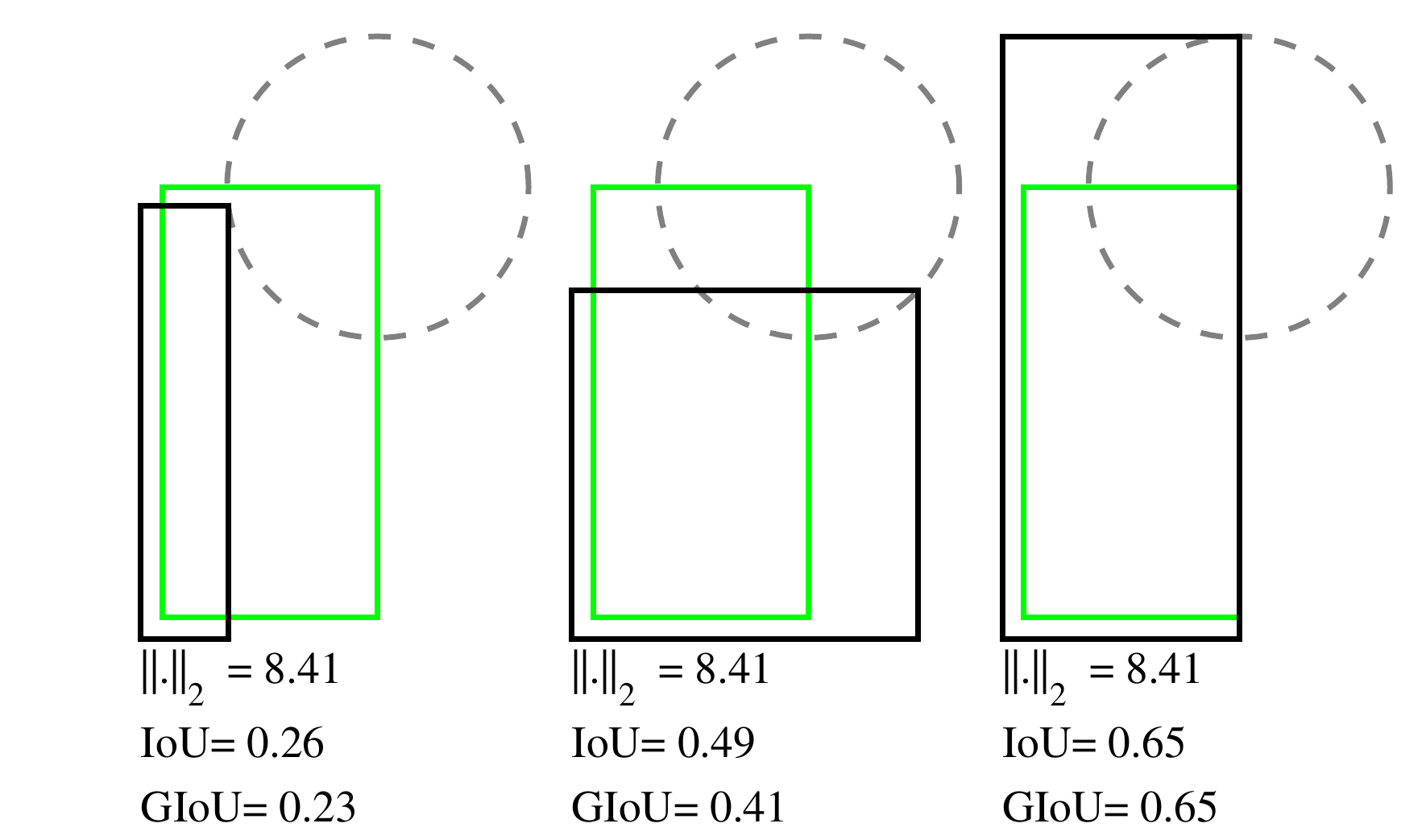}
		\centerline{(a)}\medskip
	\end{center}
	\vspace{-1.7em}
\end{minipage}
\begin{minipage}[b]{.99\linewidth}
	\begin{center}
		\includegraphics[width=.8\linewidth,trim={1.5cm 7.7cm 1.5cm 11.2cm},clip]{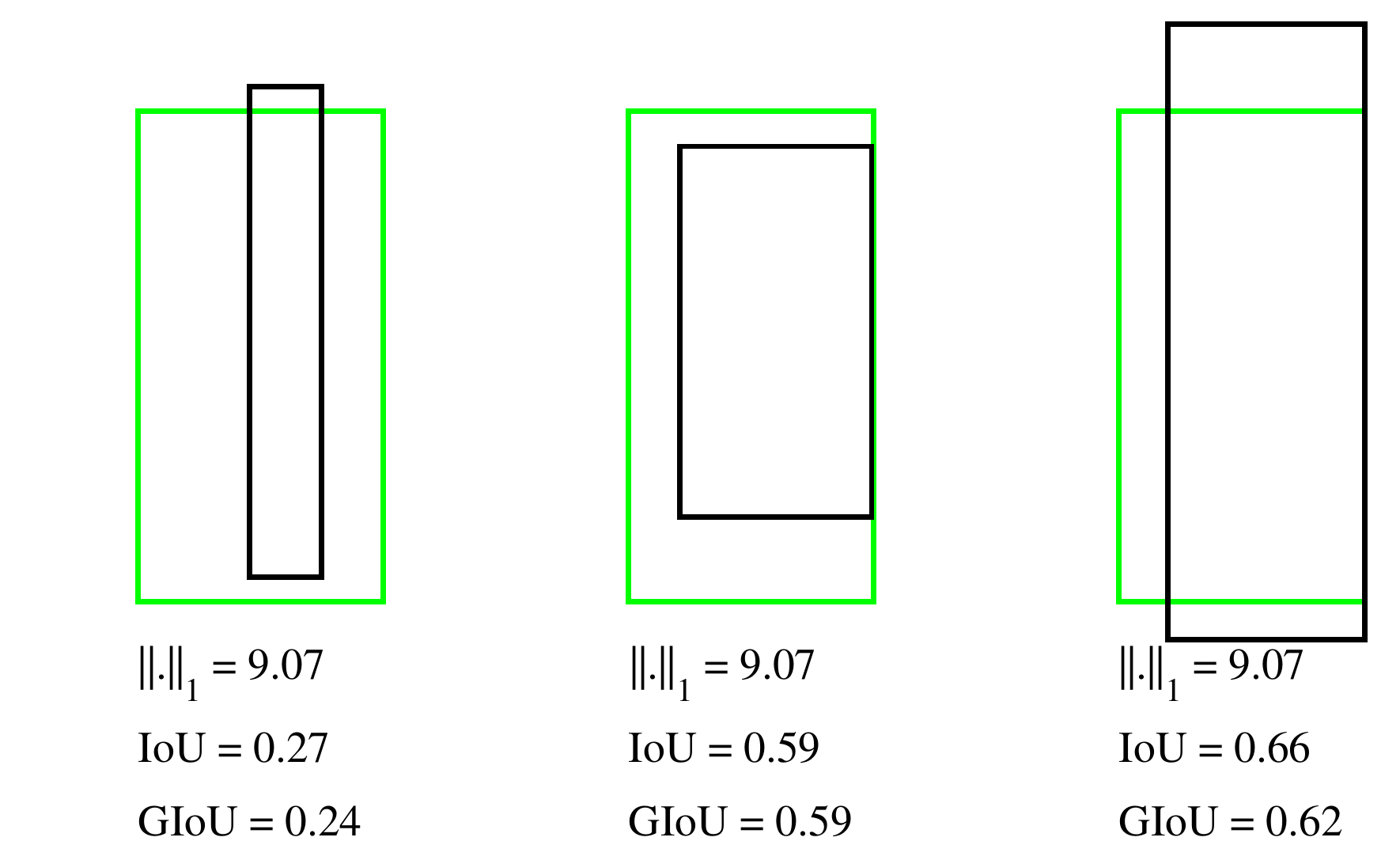}
		\centerline{(b)}\medskip
	\end{center}
\end{minipage}
	\caption{\ Two sets of examples (a) and (b) with the bounding boxes represented by (a) two corners $(x_1, y_1, x_2, y_2)$ and (b) center and size $(x_c,y_c,w,h)$. For all three cases in each set (a) $\ell_2$-norm distance, $||.||_2$, and (b) $\ell_1$-norm distance, $||.||_1$, between the representation of two rectangles are exactly same value, but their $IoU$ and $GIoU$ values are very different.}
	\label{fig:cases_dist}
\end{figure}
 
$IoU$, also known as Jaccard index, is the most commonly used metric for comparing the similarity between two arbitrary shapes. $IoU$ encodes the shape properties of the objects under comparison, \eg the widths, heights and locations of two bounding boxes, into the region property and then calculates a normalized measure that focuses on their areas (or volumes). This property makes $IoU$ \emph{invariant to the scale} of the problem under consideration. Due to this appealing property, all performance measures used to evaluate for segmentation ~\cite{Cityscapes,kitti_seg,ade20k,mscoco}, object detection~\cite{mscoco,pascal_voc}, and tracking~\cite{mot2015,vot2016} rely on this metric.

However, it can be shown that there is not a strong correlation between minimizing the commonly used losses, \eg $\ell_n$-norms, defined on parametric representation of two bounding boxes in 2D/3D and improving their $IoU$ values.
For example, consider the simple 2D scenario in \Fig~\ref{fig:cases_dist} (a), where the predicted bounding box (black rectangle), and the ground truth box (green rectangle), are represented by their top-left and bottom-right corners, \ie $(x_1,y_1,x_2,y_2)$. For simplicity, let's assume that the distance, \eg $\ell_2$-norm, between one of the corners of two boxes is fixed. Therefore any predicted bounding box where the second corner lies on a circle with a fixed radius centered on the second corner of the green rectangle (shown by a gray dashed line circle) will have exactly the same $\ell_2$-norm distance from the ground truth box; however their $IoU$ values can be significantly different (\Fig~\ref{fig:cases_dist} (a)). The same argument can be extended to any other representation and loss, \eg \Fig~\ref{fig:cases_dist} (b). It is intuitive that a good local optimum for these types of objectives may not necessarily be a local optimum for $IoU$.  Moreover, in contrast to $IoU$, $\ell_n$-norm objectives defined based on the aforementioned parametric representations are not invariant to the scale of the problem. To this end, several pairs of bounding boxes with the same level of overlap, but different scales due to \eg perspective, will have different objective values. In addition, some representations may suffer from lack of regularization between the different types of parameters used for the representation. For example, in the center and size representation, $(x_c,y_c)$ is defined on the location space while $(w,h)$ belongs to the size space. Complexity increases as more parameters are incorporated, \eg rotation, or when adding more dimensions to the problem. To alleviate some of the aforementioned problems,  state-of-the-art object detectors introduce the concept of an anchor box~\cite{faster_rcnn} as a hypothetically good initial guess. They also define a non-linear representation~\cite{yolov1, rcnn} to naively compensate for the scale changes. Even with these handcrafted changes, there is still a gap between optimizing the regression losses and $IoU$ values.

In this paper, we explore the calculation of $IoU$ between two axis aligned rectangles, or generally two axis aligned n-orthotopes, which has a straightforward analytical solution and in contrast to the prevailing belief, $IoU$ in this case can be backpropagated~\cite{iou_loss}, \ie it can be directly used as the objective function to optimize. It is therefore preferable to use $IoU$ as the objective function for 2D object detection tasks. Given the choice between optimizing a metric itself vs. a surrogate loss function, the optimal choice is the metric itself. However, $IoU$ as both a metric and a loss has a major issue: if two objects do not overlap, the $IoU$ value will be zero and will not reflect how far the two shapes are from each other. In this case of non-overlapping objects, if $IoU$ is used as a loss, its gradient will be zero and cannot be optimized.
%(ii) $IoU$ cannot properly distinguish between different alignments of two objects. More precisely, $IoU$ for two objects overlapping in several different orientations with the same intersection level will be exactly equal (\Fig~\ref{fig:cases}). Therefore, the value of the $IoU$ function does not reflect how overlap between two objects occurs. We will further elaborate on this issue in the paper.

In this paper, we will address this weakness of $IoU$ by extending the concept to non-overlapping cases. We ensure this generalization (a) follows the same definition as $IoU$, \ie encoding the shape properties of the compared objects into the region property; (b) maintains the scale invariant property of $IoU$, and (c) ensures a strong correlation with $IoU$ in the case of overlapping objects. We introduce this generalized version of $IoU$, named $GIoU$, as a new metric for comparing any two arbitrary shapes. We also provide an analytical solution for calculating $GIoU$ between two axis aligned rectangles, allowing it to be used as a loss in this case. Incorporating $GIoU$ loss into state-of-the art object detection algorithms, we consistently improve their performance on popular object detection benchmarks such as PASCAL VOC~\cite{pascal_voc} and MS COCO~\cite{mscoco} using both the standard, \ie $IoU$ based~\cite{pascal_voc,mscoco}, and the new, $GIoU$ based, performance measures.  

The main contribution of the paper is summarized as follows:
 \begin{itemize}
  \setlength{\itemsep}{1pt}
  \setlength{\parskip}{0pt}
  \setlength{\parsep}{0pt}
 \item We introduce this generalized version of $IoU$, as a new metric for comparing any two arbitrary shapes.
 \item We provide an analytical solution for using  $GIoU$ as loss between two axis-aligned rectangles or generally n-orthotopes\footnote{Extension provided in supp. material}.
 \item We incorporate $GIoU$ loss into the most popular object detection algorithms such as Faster R-CNN, Mask R-CNN and YOLO v3, and show their performance improvement on standard object detection benchmarks.
\end{itemize}

% !TEX root = main.tex

\section{Related Work}
\label{related work}

\textbf{Object detection accuracy measures:} 
Intersection over Union ($IoU$) is the defacto evaluation metric used in object detection. It is used to determine true positives and false positives in a set of predictions. When using $IoU$ as an evaluation metric an accuracy threshold must be chosen. For instance in the PASCAL VOC challenge~\cite{pascal_voc}, the widely reported detection accuracy measure, \ie mean Average Precision (mAP), is calculated based on a fixed $IoU$ threshold, \ie $0.5$. However, an arbitrary choice of the $IoU$ threshold does not fully reflect  the localization performance of different methods. Any localization accuracy higher than the threshold is treated equally. In order to make this performance measure less sensitive to the choice of $IoU$ threshold, the MS COCO Benchmark challenge~\cite{mscoco} averages  mAP across multiple $IoU$ thresholds. 

\textbf{Bounding box representations and losses:} In 2D object detection, learning bounding box parameters is crucial. Various bounding box representations and losses have been proposed in the literature. Redmon \etal in YOLO v1\cite{yolov1} propose a direct regression on the bounding box parameters with a small tweak to predict square root of the bounding box size to remedy scale sensitivity. Girshick \etal \cite{rcnn} in R-CNN parameterize the bounding box representation by predicting location and size offsets from a prior bounding box calculated using a selective search algorithm~\cite{uijlings2013selective}. To alleviate scale sensitivity of the representation, the bounding box size offsets are defined in log-space. Then, an $\ell_2$-norm objective, also known as MSE loss, is used as the objective to optimize. Later, in Fast R-CNN \cite{fast_rcnn}, Girshick proposes $\ell_1$-smooth loss to make the learning more robust against outliers. Ren \etal \cite{faster_rcnn} propose the use of a set of dense prior bounding boxes, known as anchor boxes, followed by a regression to small variations on bounding box locations and sizes. However, this makes training the bounding box scores more difficult due to significant class imbalance between positive and negative samples. To mitigate this problem, the authors later introduce focal loss~\cite{retina}, which is orthogonal to the main focus of our paper.

Most popular object detectors~\cite{yolov2,yolov3,rfcn,fpn,retina,ssd} utilize some combination of the bounding box representations and losses mentioned above. These considerable efforts have yielded significant improvement in object detection. We show there may be some opportunity for further improvement in localization with the use of $GIoU$, as their bounding box regression losses are not directly representative of the core evaluation metric, \ie $IoU$. 

\textbf{Optimizing $IoU$ using an approximate or a surrogate function:} In the semantic segmentation task, there have been some efforts to optimize $IoU$ using either an approximate function~\cite{rahman2016optimizing} or a surrogate loss~\cite{matthew2018lovasz}. Similarly, for the object detection task, recent works \cite{jiang2018acquisition, iou_loss} have attempted to directly or indirectly incorporate $IoU$ to better perform bounding box regression. However, they suffer from either an approximation or a plateau which exist in optimizing $IoU$ in non-overlapping cases. In this paper we address the weakness of $IoU$ by introducing a generalized version of $IoU$, which is directly incorporated as a loss for the object detection problem.

\section{Generalized Intersection over Union}
\label{GIOU}
Intersection over Union ($IoU$) for comparing similarity between two arbitrary shapes (volumes) $A, B\subseteq\mathbb{S}\in\mathbb{R}^n$ is attained by:
\begin{equation}
IoU = \frac{|A\cap B|}{|A\cup B|}
\end{equation}

Two appealing features, which make this similarity measure popular for evaluating many 2D/3D computer vision tasks are as follows: 
\begin{itemize}
	\item $IoU$ as a distance, \eg $\calL_{IoU} = 1- IoU$, is a metric (by mathematical definition)~\cite{kosub2016note}. It means $\calL_{IoU}$ fulfills all properties of a metric such as non-negativity, identity of indiscernibles,	symmetry and triangle inequality. 
	\item $IoU$ is invariant to the scale of the problem. This means that the similarity between two arbitrary shapes $A$ and $B$ is independent from the scale of their space $\mathbb{S}$ (the proof is provided in supp. material).   
\end{itemize}
However, $IoU$ has a major weakness: 
\begin{itemize}
	\item If $|A\cap B| = 0$, $IoU(A,B) = 0$. In this case, $IoU$ does not reflect if two shapes are in vicinity of each other or very far from each other. 
% 	\item  $IoU$ value for different alignments of two shapes is identical as long as the volume (area) of their intersection in each case is equal. Therefore, $IoU$ does not reflect how overlap between two objects occurs (\Fig~\ref{fig:cases}). 
\end{itemize}
% \begin{figure}[tb]

% \begin{minipage}[b]{.325\linewidth}
% 	\begin{center}
% 		\includegraphics[width=1.\linewidth,trim={5cm 8cm 5cm 9.3cm},clip]{figs/case4.pdf}
% 	\end{center}
% \end{minipage}
% \hfill
% \begin{minipage}[b]{.325\linewidth}
% 	\begin{center}
% 		\includegraphics[width=1.\linewidth,trim={5cm 8cm 5cm 9.3cm},clip]{figs/case5.pdf}
% 	\end{center}
% \end{minipage}
% \hfill
% \begin{minipage}[b]{.325\linewidth}
% 	\begin{center}
% 		\includegraphics[width=1.\linewidth,trim={5.cm 8cm 5.cm 9.3cm},clip]{figs/case6.pdf}
% 	\end{center}
% 	\end{minipage}

% 	\caption{Three different ways of overlap between two rectangles with the exactly same $IoU$ values, \ie $IoU = 0.33$, but different $GIoU$ values, \ie from the left to right $GIoU = 0.33, 0.24$ and $-0.1$ respectively. $GIoU$ value will be higher for the cases with better aligned orientation. }
% 	\label{fig:cases}
% \end{figure}

To address this issue, we propose a general extension to $IoU$, namely Generalized Intersection over Union $GIoU$. 

For two arbitrary convex shapes (volumes) $A, B\subseteq\mathbb{S}\in\mathbb{R}^n$, we first find the smallest convex shapes $C\subseteq\mathbb{S}\in\mathbb{R}^n$ enclosing both $A$ and $B$\footnote{Extension to non-convex cases has been provided in supp. material.}. For comparing two specific types of geometric shapes,  $C$ can be from the same type. For example, two arbitrary ellipsoids, $C$ could be the smallest ellipsoids enclosing them. Then we calculate a ratio between the volume (area) occupied by $C$ excluding $A$ and $B$ and divide by the total volume (area) occupied by $C$. This represents a normalized measure that focuses on the empty volume (area) between $A$ and $B$. Finally $GIoU$ is attained by subtracting this ratio from the $IoU$ value. The calculation of $GIoU$ is summarized in \Alg~\ref{algo:GIOU}.
\begin{algorithm}[!t]\label{algo:GIOU}	\caption{Generalized Intersection over Union}
	\small{
	\SetKwInOut{Input}{input}\SetKwInOut{Output}{output}
		\Input{Two arbitrary convex shapes: $A,B\subseteq\mathbb{S}\in\mathbb{R}^n$}
		\Output{$GIoU$}}
		For $A$ and $B$, find the smallest enclosing convex object $C$, where $C\subseteq\mathbb{S}\in\mathbb{R}^n$ \\
		$\displaystyle IoU = \frac{|A\cap B|}{|A\cup B|}$\\
		$\displaystyle GIoU = IoU - \frac{|C\backslash(A\cup B)|}{|C|}$
\end{algorithm}

$GIoU$ as a new metric has the following properties:~\footnote{Their proof has been provided in supp. material. } 

\begin{enumerate}
    \item Similar to $IoU$, $GIoU$ as a distance, \eg $\calL_{GIoU} = 1- GIoU$, holding all properties of a metric such as non-negativity, identity of indiscernibles,	symmetry and triangle inequality.
	\item Similar to $IoU$, $GIoU$ is invariant to the scale of the problem.
	\item $GIoU$ is always a lower bound for $IoU$, \ie $\forall A, B\subseteq\mathbb{S}$  $GIoU(A,B)\leq IoU(A,B)$, and this lower bound becomes tighter when $A$ and $B$ have a stronger shape similarity and proximity, \ie  $\lim_{A\to B} GIoU(A,B) = IoU(A,B)$. 
	\item $\forall A, B\subseteq\mathbb{S}$, $  0\leq IoU(A,B)\leq 1$, but $GIoU$ has a symmetric range, \ie $\forall A, B\subseteq\mathbb{S}$, $   -1\leq GIoU(A,B) \leq 1$. 
	\begin{enumerate}[I)]
	\item Similar to $IoU$, the value $1$ occurs only when two objects overlay perfectly, \ie if $|A\cup B| = |A\cap B|$, then $GIoU = IoU = 1$ 
	\item $GIoU$ value asymptotically converges to -1 when the ratio between occupying regions of two shapes, $|A\cup B|$, and the volume (area) of the enclosing shape $|C|$ tends to zero, \ie $\displaystyle \lim_{\frac{|A\cup B|}{|C|}\to 0} GIoU(A,B) = -1$ . 
	\end{enumerate}
% 	\item In contrast to $IoU$, $GIoU$ does not only focus on overlapping area. The empty space between two \emph{symmetrical} shapes $A$ and $B$ in the enclosing shape $C$ increases when $A$ and $B$ are not well aligned with respect to each other (\Fig~\ref{fig:cases}). Therefore, the value of $GIoU$ can better reflect how overlap between two symmetrical objects occurs. 
\end{enumerate}

% The reason we care about the last property is that a metric that reflects changes in orientation between two shapes allows differentiation between results that would otherwise be identical.

In summary, this generalization keeps the major properties of $IoU$ while rectifying its weakness. Therefore, $GIoU$ can be a proper substitute for $IoU$ in all performance measures used in 2D/3D computer vision tasks. In this paper, we only focus on 2D object detection where we can easily derive an analytical solution for $GIoU$ to apply it as both metric and loss. The extension to non-axis aligned 3D cases is left as future work. 

\subsection{GIoU as Loss for Bounding Box Regression}
So far, we introduced $GIoU$ as a metric for any two arbitrary shapes. However as is the case with $IoU$, there is no analytical solution for calculating intersection between two arbitrary shapes and/or for finding the smallest enclosing convex object for them.  

Fortunately, for the 2D object detection task where the task is to compare two axis aligned bounding boxes, we can show that $GIoU$ has a straightforward solution. In this case, the intersection and the smallest enclosing objects both have rectangular shapes. It can be shown that the coordinates of their vertices are simply the coordinates of one of the two bounding boxes being compared, which can be attained by comparing each vertices' coordinates using \emph{min} and \emph{max} functions. To check if two bounding boxes overlap, a condition must also be checked. Therefore, we have an exact solution to calculate $IoU$ and $GIoU$. 

\begin{figure}[b]
\vspace{-1em}
  \centering
    \includegraphics[width=1.1\linewidth,trim={1cm 8.9cm 0cm 9.3cm},clip]{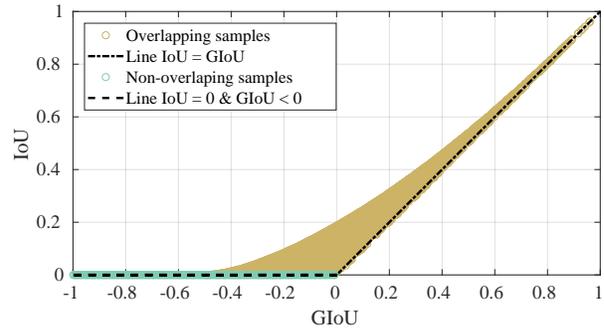}
	\caption{\small Correlation between GIoU and IOU for overlapping and non-overlapping samples.}
	\label{fig:GIoUvsIoU}
	\vspace{-1em}
\end{figure}

Since back-propagating \emph{min}, \emph{max} and piece-wise linear functions, \eg \emph{Relu}, are feasible, it can be shown that every component in \Alg~\ref{algo:GIOU loss} has a well-behaved derivative. Therefore, $IoU$ or $GIoU$ can be directly used as a loss, \ie $\calL_{IoU}$ or $\calL_{GIoU}$, for optimizing deep neural network based object detectors. In this case, we are directly optimizing a metric as loss, which is an optimal choice for the metric. However, in all non-overlapping cases, $IoU$ has zero gradient, which affects both training quality and convergence rate. $GIoU$, in contrast, has a gradient in all possible cases, including non-overlapping situations. In addition, using property 3, we show that $GIoU$ has a strong correlation with $IoU$, especially in high $IoU$ values. We also demonstrate this correlation qualitatively in \Fig~\ref{fig:GIoUvsIoU} by taking over 10K random samples from the coordinates of two 2D rectangles. In \Fig~\ref{fig:GIoUvsIoU}, we also observe that in the case of low overlap, \eg $IoU\leq 0.2$ and $GIoU\leq 0.2$, $GIoU$ has the opportunity to change more dramatically compared to $IoU$. To this end, $GIoU$ can potentially have a steeper gradient in any possible state in these cases compared to $IoU$. Therefore, optimizing $GIoU$ as loss, $\calL_{GIoU}$ can be a better choice compared to $\calL_{IoU}$, no matter which $IoU$-based performance measure is ultimately used. Our experimental results verify this claim. 
\begin{algorithm}[!tb]\label{algo:GIOU loss}
\caption{$IoU$ and $GIoU$ as bounding box losses}
	\small{
	\SetKwInOut{Input}{input}\SetKwInOut{Output}{output}
		\Input{Predicted $B^p$ and ground truth $B^g$ bounding box coordinates: $B^p = (x^p_1,y^p_1,x^p_2,y^p_2) $,$\quad B^g = (x^g_1,y^g_1,x^g_2,y^g_2)$.}
		\Output{$\calL_{IoU}$, $\calL_{GIoU}$.}}
		For the predicted box $B^p$, ensuring  $x^p_2>x^p_1$ and $y^p_2>y^p_1$:
		$\htx^p_1 = \min(x^p_1,x^p_2)$, $\quad \htx^p_2 = \max(x^p_1,x^p_2)$, $\quad\hty^p_1 = \min(y^p_1,y^p_2)$, $\quad\hty^p_2 = \max(y^p_1,y^p_2)$.\\
		Calculating area of $B^g$: $A^g = (x^g_2 - x^g_1)\times(y^g_2 - y^g_1)$.\\
		Calculating area of $B^p$: $A^p = (\htx^p_2 - \htx^p_1)\times(\hty^p_2 - \hty^p_1)$.\\
		Calculating intersection $\calI$ between $B^p$ and $B^g$:
		$x^{\calI}_1 = \max(\htx^p_1,x^g_1)$, $\quad x^{\calI}_2 = \min(\htx^p_2,x^g_2)$, $\quad y^{\calI}_1 = \max(\hty^p_1,y^g_1)$, $\quad y^{\calI}_2 = \min(\hty^p_2,y^g_2)$, $
		    \calI = \begin{cases} 
    (x^{\calI}_2 - x^{\calI}_1)\times(y^{\calI}_2 - y^{\calI}_1) & \text{if}\quad x^{\calI}_2 > x^{\calI}_1, y^{\calI}_2 > y^{\calI}_1\\
    0 & \text{otherwise.}
  \end{cases}$\\
		Finding the coordinate of smallest enclosing box $B^c$: $x^{c}_1 = \min(\htx^p_1,x^g_1)$, $\quad x^{c}_2 = \max(\htx^p_2,x^g_2)$, $\quad y^{c}_1 = \min(\hty^p_1,y^g_1)$, $\quad y^{c}_2 = \max(\hty^p_2,y^g_2)$. \\
		Calculating area of $B^c$: $A^c = (x^c_2 - x^c_1)\times(y^c_2 - y^c_1)$.\\
		$\displaystyle IoU = \frac{\calI}{\calU}$, where  $\calU = A^p+A^g-\calI$.\\
		$\displaystyle GIoU = IoU - \frac{A^c-\calU}{A^c}$.\\
		$\calL_{IoU} = 1- IoU$, 
		$\quad\calL_{GIoU} = 1- GIoU$.
\end{algorithm}

\textbf{Loss Stability:} We also investigate if there exist any extreme cases which make the loss unstable/undefined given any value for the predicted outputs. 

Considering the ground truth bounding box, $B^g$ is a rectangle with area bigger than zero, \ie $A^g>0$. \Alg~\ref{algo:GIOU loss} ($\mathbf{1}$) and the Conditions in \Alg~\ref{algo:GIOU loss} ($\mathbf{4}$) respectively ensure the predicted area $A^p$ and intersection $\calI$ are non-negative values, \ie $A^p\geq0$ and $\calI\geq0$  $\forall B^p\in \mathbb{R}^4$. Therefore union $\calU>0$ for any predicted value of $B^p = (x^p_1,y^p_1,x^p_2,y^p_2)\in\mathbb{R}^4$. This ensures that the denominator in $IoU$ cannot be zero for any predicted value of outputs. Moreover, for any values of $B^p=(x^p_1,y^p_1,x^p_2,y^p_2)\in \mathbb{R}^4$, union is always bigger than intersection, \ie $\calU\geq\calI$.  Consequently, $\calL_{IoU}$ is always bounded, \ie $0\leq \calL_{IoU}\leq 1$   $\forall B^p\in \mathbb{R}^4$. 

To check the stability of $\calL_{GIoU}$, the extra term, \ie $\frac{A^c-\calU}{A^c}$, should always be a defined and bounded value. It can be easily perceived that the smallest enclosing box $B^c$ cannot be smaller than $B^g$ for all predicted values. Therefore the denominator in $\frac{A^c-\calU}{A^c}$ is always a positive non-zero value, because $A^c\geq A^g$ $\forall B^p\in \mathbb{R}^4$ and $A^g\geq 0$. Moreover, the area of the smallest enclosing box cannot be smaller than union for any value of predictions, \ie $A^c\geq\calU$  $\forall B^p\in \mathbb{R}^4$. Therefore, the extra term in $GIoU$ is positive and bounded.  Consequently, $\calL_{GIoU}$ is always bounded, \ie $0\leq \calL_{GIoU}\leq 2$   $\forall B^p\in \mathbb{R}^4$.   

\textbf{$\calL_{GIoU}$ behaviour when IoU = 0:} 
For $GIoU$ loss, we have $\calL_{GIoU} = 1 - GIoU = 1+\frac{A^c-\calU}{A^c}-IoU$. In the case when $B^g$ and $B^p$ do not overlap, \ie $\calI = 0$ and $IoU = 0$, $GIoU$ loss simplifies to $\calL_{GIoU} = 1+\frac{A^c-\calU}{A^c}=2-\frac{\calU}{A^c}$. In this case, by minimizing $\calL_{GIoU}$, we actually maximize the term $\frac{\calU}{A^c}$. This term is a normalized measure between 0 and 1, \ie $0\leq\frac{\calU}{A^c}\leq 1$, and is maximized when the area of the smallest enclosing box $A^c$ is minimized while the union $\calU = A^g + A^p$, or more precisely the area of predicted bounding box $A^p$, is maximized. To accomplish this, the vertices of the predicted bounding box $B^p$ should move in a direction that encourages $B^g$ and $B^p$ to overlap, making $IoU\neq0$.
% !TEX root = main.tex

\section{Experimental Results}
\label{results}
We evaluate our new bounding box regression loss $\calL_{GIoU}$ by incorporating it into the most popular 2D object detectors such as Faster R-CNN~\cite{faster_rcnn}, Mask R-CNN~\cite{he2017mask} and YOLO v3~\cite{yolov3}. To this end, we replace their default regression losses with $\calL_{GIoU}$, \ie we replace $\ell_1$-smooth in Faster /Mask-RCNN~\cite{faster_rcnn,he2017mask} and MSE in YOLO v3~\cite{yolov3}. We also compare the baseline losses against $\calL_{IoU}$\footnote{All source codes including the evaluation scripts, the training codes, trained models and all loss implementations in PyTorch, TensorFlow and darknet are available at: https://giou.stanford.edu.}.
\\\\
\textbf{Dataset.}
We train all detection baselines and report all the results on two standard object detection benchmarks, \ie the PASCAL VOC~\cite{pascal_voc} and the Microsoft Common Objects in Context (MS COCO)~\cite{mscoco} challenges. The details of their training protocol and their evaluation have been provided in their own sections.   
\\
\emph{PASCAL VOC 2007:} The Pascal Visual Object Classes
(VOC)~\cite{pascal_voc} benchmark is one of the most widely used
datasets for classification, object detection and semantic segmentation. It consists of 9963 images with a 50/50 split for training and test, where objects from 20 pre-defined categories have been annotated with bounding boxes.
\\
\emph{MS COCO:} Another popular benchmark for image
captioning, recognition, detection and segmentation is the more recent Microsoft Common Objects in Context (MS-COCO)~\cite{mscoco}. The COCO dataset consists of over 200,000 images across train, validation and test sets with over 500,000 annotated object instances from 80 categories. 
\\\\
\textbf{Evaluation protocol.} In this paper, we adopt the same performance measure as the MS COCO 2018 Challenge~\cite{mscoco} to report all our results. This includes the calculation of mean Average precision (mAP) over different class labels for a specific value of $IoU$ threshold in order to determine true positives and false positives. The main performance measure used in this benchmark is shown by \textbf{AP}, which is averaging mAP across different value of $IoU$ thresholds, \ie $IoU = \{.5, .55, \cdots, .95\}$. Additionally, we modify this evaluation script to use $GIoU$ instead of $IoU$ as a metric to decide about true positives and false positives. Therefore, we report another value for \textbf{AP} by averaging mAP across different values of $GIoU$ thresholds, $GIoU = \{.5, .55, \cdots, .95\}$.  We also report the mAP value for $IoU$ and $GIoU$ thresholds equal to $0.75$, shown as \textbf{AP75} in the tables.

All detection baselines have also been evaluated using the test set of the MS COCO 2018 dataset, where the annotations are not accessible for the evaluation. Therefore in this case, we are only able to report results using the standard performance measure, \ie $IoU$.

\subsection{YOLO v3}
\textbf{Training protocol.} We used the original Darknet implementation of YOLO v3 released by the authors~\footnote{Available at: https://pjreddie.com/darknet/yolo/}. For baseline results (training using MSE loss), we used DarkNet-608 as backbone network architecture in all experiments and followed exactly their training protocol using the reported default parameters and the number of iteration on each benchmark. To train YOLO v3 using $IoU$ and $GIoU$ losses, we simply replace the bounding box regression MSE loss with $\calL_{IoU}$ and $\calL_{GIoU}$ losses explained in \Alg~\ref{algo:GIOU loss}. Considering the additional MSE loss on classification and since we replace an unbounded distance loss such as MSE distance with a bounded distance, \eg $\calL_{IoU}$ or $\calL_{GIoU}$, we need to regularize the new bounding box regression against the classification loss. However, we performed a very minimal effort to regularize these new regression losses against the MSE classification loss.

\textbf{PASCAL VOC 2007.} Following the original code's training protocol, we trained the network using each loss on both training and validation set of the dataset up to $50K$ iterations. Their performance using the best network model for each loss has been evaluated using the PASCAL VOC 2007 test and the results have been reported in \Tab~\ref{table:yolo-voc}.

Considering both standard $IoU$ based and new $GIoU$ based performance measures, the results in \Tab~\ref{table:yolo-voc} show that training YOLO v3 using $\calL_{GIoU}$ as regression loss can considerably improve its performance compared to its own regression loss (MSE). Moreover, incorporating $\calL_{IoU}$ as regression loss can slightly improve the performance of YOLO v3 on this benchmark. However, the improvement is inferior compared to the case where it is trained by $\calL_{GIoU}$.

\begin{table}[!tb]\footnotesize
\centering
   \caption{\footnotesize Comparison between the performance of \textbf{YOLO v3}~\cite{yolov3} trained using its own loss (MSE) as well as $\calL_{IoU}$ and $\calL_{GIoU}$ losses. The results are reported on the \textbf{test set of PASCAL VOC 2007.}}
 \begin{tabular}{c c  c c c c} 
 \hline 
\raisebox{-1.5ex}{Loss \big/ Evaluation} & \multicolumn{2}{c}{\raisebox{-0.5ex}{AP}} &&\multicolumn{2}{c}{\raisebox{-0.5ex}{AP75}}\\ [0.5ex] 
\cline{2-3}\cline{5-6}
 & \raisebox{-0.2ex}{\textbf{IoU}} & \raisebox{-0.2ex}{\textbf{GIoU}} &&\raisebox{-0.2ex}{\textbf{IoU}} & \raisebox{-0.2ex}{\textbf{GIoU}} \\ 
 \hline
 \hline
MSE~\cite{yolov3} & .461 & .451 && .486 & .467 \\
 \hline
 \hline
  $\calL_{IoU}$ & .466 & .460 && .504 & .498 \\
  Relative improv \% & 1.08\% & 2.02\% && 3.70\% & 6.64\% \\
 \hline
  $\calL_{GIoU}$ & \textbf{.477} & \textbf{.469} && \textbf{.513} & \textbf{.499} \\
  Relative improv \% & \textbf{3.45\%} & \textbf{4.08\%} && \textbf{5.56\%} & \textbf{6.85\%} \\
  \hline
 \end{tabular}
 \label{table:yolo-voc}
\end{table}

\textbf{MS COCO.} Following the original code's training protocol, we trained YOLO v3 using each loss on both the training set and 88\% of the validation set of MS COCO 2014 up to $502k$ iterations. Then we evaluated the results using the remaining 12\% of the validation set and reported the results in \Tab~\ref{table:yolo-cocoval}.
We also compared them on the MS COCO 2018 Challenge by submitting the results to the COCO server. All results using the $IoU$ based performance measure are reported in \Tab~\ref{table:yolo-cocotest}.
\begin{figure}[!b]
\begin{minipage}[b]{.49\linewidth}
	\begin{center}
	    \includegraphics[width=1.1\linewidth,trim={1.2cm 6.5cm 1.2cm 7cm},clip]{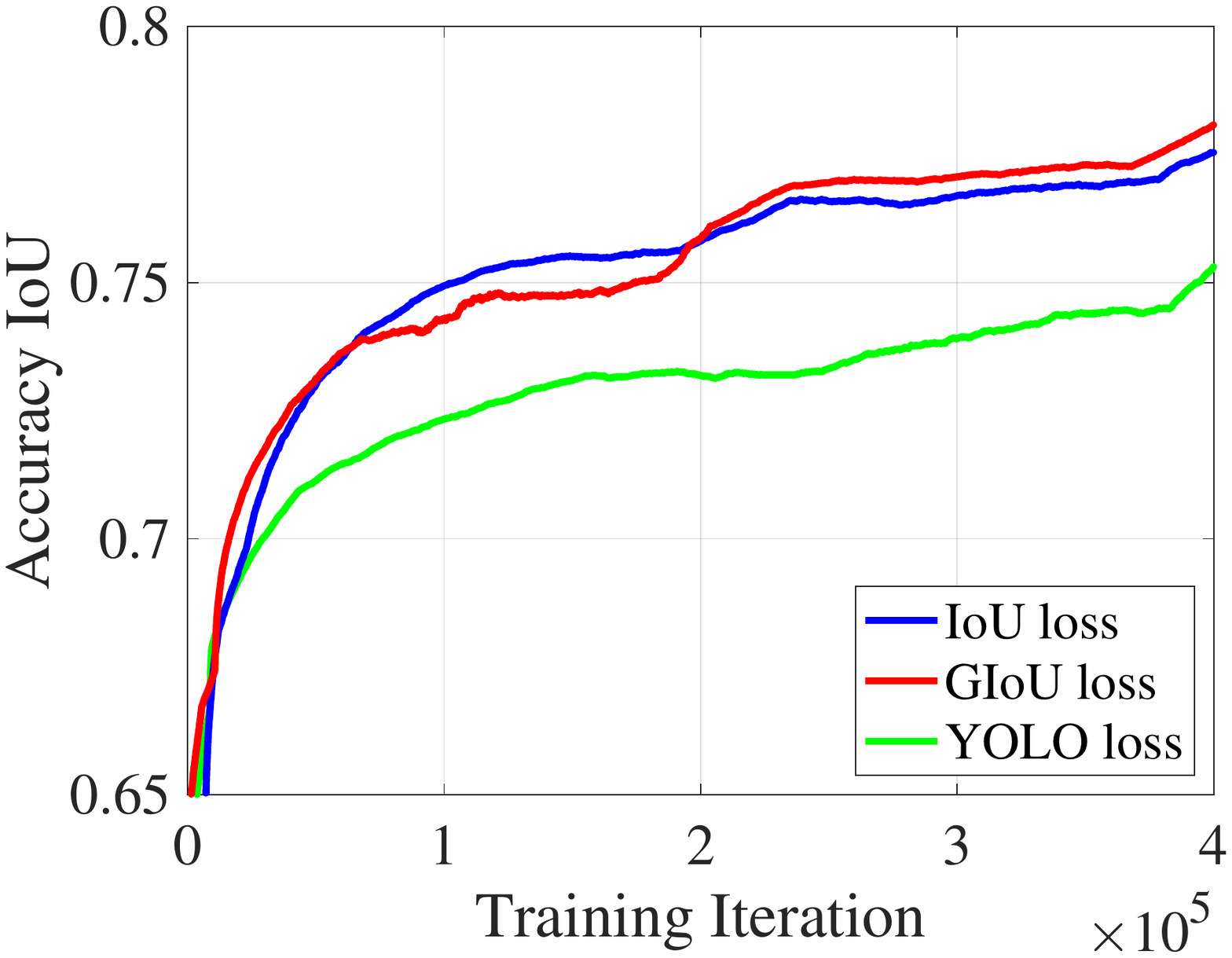}
	    \centering (a)
	\end{center}
\end{minipage}
\hfill
\begin{minipage}[b]{.49\linewidth}
	\begin{center}
	    \includegraphics[width=1.1\linewidth,trim={1.2cm 6.5cm 1.2cm 7cm},clip]{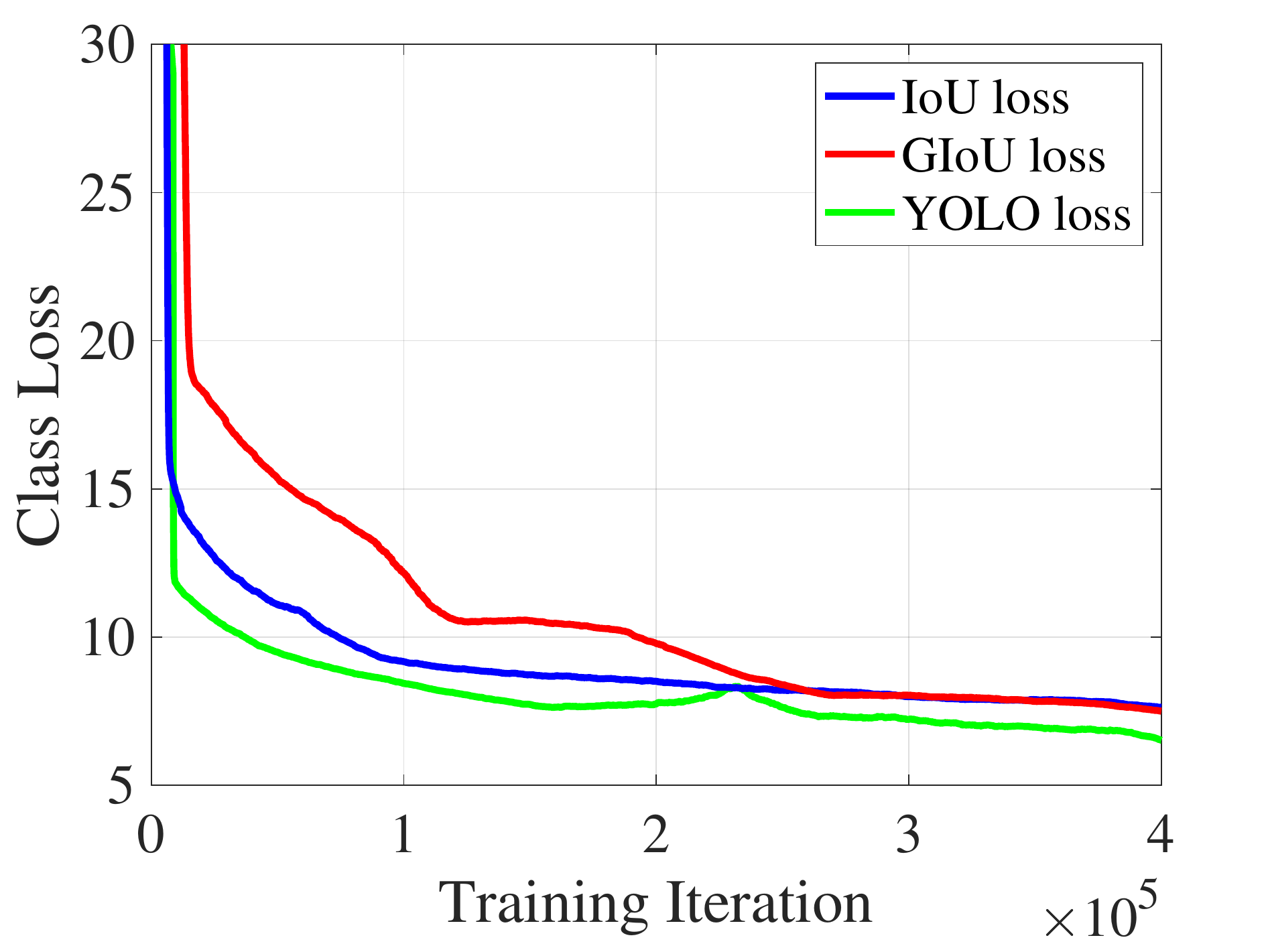}
	    \centering (b)
	\end{center}
\end{minipage}
	\caption{The classification loss and accuracy (average $IoU$) against training iterations when YOLO v3~\cite{yolov3} was trained using its standard (MSE) loss as well as $\calL_{IoU}$ and $\calL_{GIoU}$ losses.}
	\label{fig:yolo-training}
\end{figure}
Similar to the PASCAL VOC experiment, the results show consistent improvement in performance for YOLO v3 when it is trained using $\calL_{GIoU}$ as regression loss. We have also investigated how each component, \ie bounding box regression and classification losses, contribute to the final AP performance measure. We believe the localization accuracy for YOLO v3 significantly improves when $\calL_{GIoU}$ loss is used (\Fig~\ref{fig:yolo-training} (a)). However, with the current naive tuning of regularization parameters, balancing bounding box loss vs. classification loss, the classification scores may not be optimal, compared to the baseline (\Fig~\ref{fig:yolo-training} (b)). Since AP based performance measure is considerably affected by small classification error, we believe the results can be further improved with a better search for regularization parameters.

% \begin{table}[!tb]\footnotesize
% \centering
%   \caption{\footnotesize Comparison between the performance of \textbf{YOLO v3}~\cite{yolov3} trained using its own loss (MSE) as well as $\calL_{IoU}$ and $\calL_{GIoU}$ losses. The results are reported on 5K of the \textbf{2014 validation set of MS COCO}.}
%  \begin{tabular}{c c  c c c c}
%  \hline
% \raisebox{-1.5ex}{Loss \big/ Evaluation} & \multicolumn{2}{c}{\raisebox{-0.5ex}{AP}} &&\multicolumn{2}{c}{\raisebox{-0.5ex}{AP75}}\\ [0.5ex]
% \cline{2-3}\cline{5-6}
%  & \raisebox{-0.2ex}{\textbf{IoU}} & \raisebox{-0.2ex}{\textbf{GIoU}} &&\raisebox{-0.2ex}{\textbf{IoU}} & \raisebox{-0.2ex}{\textbf{GIoU}} \\
%  \hline
%  \hline
% MSE~\cite{yolov3}& .283 & .312 && .289 & .330 \\
%  \hline
%  \hline
%  $\calL_{IoU}$ & .292 & .320 && .312 & .346\\
%   Relative improv \% & 3.18\% & 2.56\% && 7.96\% & 4.85\% \\
%  \hline
%  $\calL_{GIoU}$ & \textbf{.301} & \textbf{.332} && \textbf{.325} & \textbf{.359} \\
%   Relative improv \% & \textbf{6.36\%} & \textbf{6.41\%} && \textbf{12.46\%} & \textbf{8.79\%} \\
%  \hline
%  \end{tabular}
%   \label{table:yolo-cocoval}
% \end{table}

% We continued training YOLO v3 with MSE, $IoU$ and $GIoU$ losses for 50k iterations and found slight improvements across all losses on both the validation and test set of the MS COCO dataset.
\begin{table}[!tb]\footnotesize
\centering
%   \caption{\footnotesize Comparison between the performance of YOLO v3~\cite{yolov3} trained using its own loss (MSE), and using $IoU$ and $GIoU$ losses. The results are reported on 5K of the \textbf{2014 validation set}}
  \caption{\footnotesize Comparison between the performance of \textbf{YOLO v3}~\cite{yolov3} trained using its own loss (MSE) as well as $\calL_{IoU}$ and $\calL_{GIoU}$ losses. The results are reported on 5K of the \textbf{2014 validation set of MS COCO}.}
 \begin{tabular}{c c  c c c c}
 \hline
\raisebox{-1.5ex}{Loss \big/ Evaluation} & \multicolumn{2}{c}{\raisebox{-0.5ex}{AP}} &&\multicolumn{2}{c}{\raisebox{-0.5ex}{AP75}}\\ [0.5ex]
\cline{2-3}\cline{5-6}
 & \raisebox{-0.2ex}{\textbf{IoU}} & \raisebox{-0.2ex}{\textbf{GIoU}} &&\raisebox{-0.2ex}{\textbf{IoU}} & \raisebox{-0.2ex}{\textbf{GIoU}} \\
 \hline
 \hline
MSE~\cite{yolov3} & 0.314 & 0.302 && 0.329 & 0.317 \\
 \hline
 \hline
 $\calL_{IoU}$ & 0.322 & 0.313 && 0.345 & 0.335\\
 Relative improv \% & 2.55\% & 3.64\% && 4.86\% & 5.68\% \\
 \hline
 $\calL_{GIoU}$ & \textbf{0.335} & \textbf{0.325} && \textbf{0.359} & \textbf{0.348} \\
  Relative improv \% & \textbf{6.69\%} & \textbf{7.62\%} && \textbf{9.12\%} & \textbf{9.78\%} \\
 \hline
 \end{tabular}
 \label{table:yolo-cocoval}
\end{table}

% \begin{table}[!tb]\footnotesize
% \centering
%     \caption{\footnotesize Comparison between the performance of \textbf{YOLO v3}~\cite{yolov3} trained using its own loss (MSE) as well as using $\calL_{IoU}$ and $\calL_{GIoU}$ losses. The results are reported on the \textbf{test set of MS COCO 2018}.}
%  \begin{tabular}{c c c c}
%  \hline
%  Loss \big/ Evaluation & AP & AP75\\ [0.5ex]
%  \hline
%  \hline
%  MSE~\cite{yolov3} & .311 & .330 \\
%  \hline
%  \hline
%  $\calL_{IoU}$ & .312 & .338 \\
%  Relative improv \% & 0.32\% & 2.37\% \\
%  \hline
%  $\calL_{GIoU}$ & \textbf{.329} & \textbf{.356} \\
%  Relative improv \% & \textbf{5.47\%} & \textbf{7.30\%} \\
%  \hline
%  \end{tabular}
%  \label{table:yolo-cocotest}
% \end{table}

\begin{table}[!tb]\footnotesize
\centering
%    \caption{\footnotesize Comparison between the performance of YOLO v3~\cite{yolov3} trained using its own loss (MSE) as well as using $\calL_{IoU}$ and $\calL_{GIoU}$ losses. The results are reported on the \textbf{test set of MS COCO 2018}.}
    \caption{\footnotesize Comparison between the performance of \textbf{YOLO v3}~\cite{yolov3} trained using its own loss (MSE) as well as using $\calL_{IoU}$ and $\calL_{GIoU}$ losses. The results are reported on the \textbf{test set of MS COCO 2018}.}
 \begin{tabular}{c c c c}
 \hline
 Loss \big/ Evaluation & AP & AP75\\ [0.5ex]
 \hline
 \hline
 MSE~\cite{yolov3} & .314 & .333 \\
 \hline
 \hline
% PRETRAINED~\cite{yolov3} & 0.33 & 0.343 & 0.577 \\
% \hline
% \hline
 $\calL_{IoU}$ & .321 & .348 \\
 Relative improv \% & 2.18\% & 4.31\% \\
 \hline
 $\calL_{GIoU}$ & \textbf{.333} & \textbf{.362} \\
 Relative improv \% & \textbf{5.71\%} & \textbf{8.01\%} \\
 \hline
 \end{tabular}
 \label{table:yolo-cocotest}
\end{table}

\subsection{Faster R-CNN and Mask R-CNN}

\textbf{Training protocol.} We used the latest PyTorch implementations of Faster R-CNN~\cite{faster_rcnn} and Mask R-CNN~\cite{he2017mask}~\footnote{https://github.com/roytseng-tw/Detectron.pytorch}, released by Facebook research. This code is analogous to the original Caffe2 implementation~\footnote{https://github.com/facebookresearch/Detectron}. For baseline results (trained using $\ell_1$-smooth), we used ResNet-50 the backbone network architecture for both Faster R-CNN and Mask R-CNN in all experiments and followed their training protocol using the reported default parameters and the number of iteration on each benchmark. To train Faster R-CNN and Mask R-CNN using $IoU$ and $GIoU$ losses, we replaced their $\ell_1$-smooth loss in the final bounding box refinement stage with $\calL_{IoU}$ and $\calL_{GIoU}$ losses explained in \Alg~\ref{algo:GIOU loss}. Similar to the YOLO v3 experiment, we undertook minimal effort to regularize the new regression loss against the other losses such as classification and segmentation losses. We simply multiplied $\calL_{IoU}$ and $\calL_{GIoU}$ losses by a factor of $10$ for all experiments.

\textbf{PASCAL VOC 2007.} Since there is no instance mask annotation available in this dataset, we did not evaluate Mask R-CNN on this dataset. Therefore, we only trained Faster R-CNN using the aforementioned bounding box regression losses on the training set of the dataset for 20k iterations. Then, we searched for the best-performing model on the validation set over different parameters such as the number of training iterations and bounding box regression loss regularizer. The final results on the test set of the dataset have been reported in \Tab~\ref{table:faster-rcnn-voc}. 

\begin{figure}[!tb]
	\begin{center}
	    \includegraphics[width=0.8\linewidth,trim={1.2cm 6.5cm 1.2cm 7cm},clip]{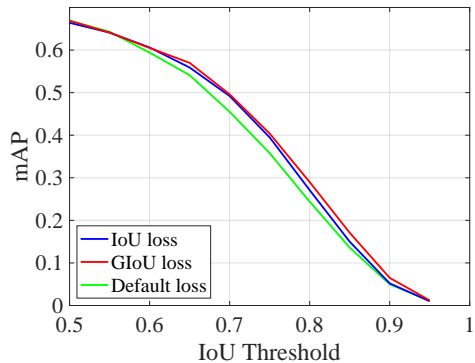}
	\end{center}
		\vspace{-1em}
	\caption{mAP value against different $IoU$ thresholds, \ie $.5 \leq IoU\leq.95$, for \textbf{Faster R-CNN} trained using $\ell_1$-smooth (green), $\calL_{IoU}$ (blue) and $\calL_{GIoU}$ (red) losses.}
	%\vspace{-1em}
	\label{fig:Faster mAP}
\end{figure}

According to both standard $IoU$ based and new $GIoU$ based performance measure, the results in \Tab~\ref{table:faster-rcnn-voc} show that training Faster R-CNN using $\calL_{GIoU}$ as the bounding box regression loss can consistently improve its performance compared to its own regression loss ($\ell_1$-smooth). Moreover, incorporating $\calL_{IoU}$ as the regression loss can slightly improve the performance of Faster R-CNN on this benchmark. The improvement is inferior compared to the case where it is trained using $\calL_{GIoU}$, see Fig.~\ref{fig:Faster mAP}, where we visualized different values of mAP against different value of $IoU$ thresholds, \ie $.5 \leq IoU\leq.95$.

\begin{table}[!b]\footnotesize
\centering
   \caption{\footnotesize Comparison between the performance of \textbf{Faster R-CNN}~\cite{faster_rcnn} trained using its own loss ($\ell_1$-smooth) as well as $\calL_{IoU}$ and $\calL_{GIoU}$ losses. The results are reported on the \textbf{test set of PASCAL VOC 2007}.}
 \begin{tabular}{c c  c c c c} 
 \hline 
\raisebox{-1.5ex}{Loss \big/ Evaluation} & \multicolumn{2}{c}{\raisebox{-0.5ex}{AP}} &&\multicolumn{2}{c}{\raisebox{-0.5ex}{AP75}}\\ [0.5ex] 
\cline{2-3}\cline{5-6}
 & \raisebox{-0.2ex}{\textbf{IoU}} & \raisebox{-0.2ex}{\textbf{GIoU}} &&\raisebox{-0.2ex}{\textbf{IoU}} & \raisebox{-0.2ex}{\textbf{GIoU}} \\ 
 \hline
 \hline
$\ell_1$-smooth~\cite{faster_rcnn} & .370 & .361 && .358 & .346 \\
 \hline
 \hline
 $\calL_{IoU}$ & .384 & .375 && .395 & .382 \\
Relative improv. \% & 3.78\% & 3.88\% && 10.34\% & 10.40\% \\
 \hline
  $\calL_{GIoU}$ & \textbf{.392} & \textbf{.382} && \textbf{.404} & \textbf{.395} \\
Relative improv. \% & \textbf{5.95\%} & \textbf{5.82\%} && \textbf{12.85\%} & \textbf{14.16\%} \\
  \hline
 \end{tabular}
 \label{table:faster-rcnn-voc}
 \vspace{-1em}
\end{table}

\textbf{MS COCO.} Similarly, we trained both Faster R-CNN and Mask R-CNN using each of the aforementioned bounding box regression losses on the MS COCO 2018 training dataset for 95K iterations. The results for the best model on the validation set of MS COCO 2018 for Faster R-CNN and Mask R-CNN have been reported in Tables~\ref{table:faster-rcnn-cocoval} and~\ref{table:mask-rcnn-cocoval} respectively.
We have also compared them on the MS COCO 2018 Challenge by submitting their results to the COCO server. All results using the $IoU$ based performance measure are also reported in Tables~\ref{table:faster-rcnn-cocotest} and ~\ref{table:mask-rcnn-cocotest}. 

Similar to the above experiments, detection accuracy improves by using $\calL_{GIoU}$ as regression loss over $\ell_1$-smooth~\cite{faster_rcnn,he2017mask}. However, the amount of improvement between different losses is less than previous experiments. This may be due to several factors. First, the detection anchor boxes on Faster R-CNN~\cite{faster_rcnn} and Mask R-CNN~\cite{he2017mask} are more dense than YOLO v3~\cite{yolov3}, resulting in less frequent scenarios where $\calL_{GIoU}$ has an advantage over $\calL_{IoU}$ such as non-overlapping bounding boxes. Second, the bounding box regularization parameter has been naively tuned on PASCAL VOC, leading to sub-optimal result on MS COCO~\cite{mscoco}.

\begin{table}[!tb]\footnotesize
\centering
   \caption{\footnotesize Comparison between the performance of \textbf{Faster R-CNN}~\cite{faster_rcnn} trained using its own loss ($\ell_1$-smooth) as well as $\calL_{IoU}$ and $\calL_{GIoU}$ losses. The results are reported on the \textbf{validation set of MS COCO 2018.}}
 \begin{tabular}{c c  c c c c} 
 \hline 
\raisebox{-1.5ex}{Loss \big/ Evaluation} & \multicolumn{2}{c}{\raisebox{-0.5ex}{AP}} &&\multicolumn{2}{c}{\raisebox{-0.5ex}{AP75}}\\ [0.5ex] 
\cline{2-3}\cline{5-6}
 & \raisebox{-0.2ex}{\textbf{IoU}} & \raisebox{-0.2ex}{\textbf{GIoU}} &&\raisebox{-0.2ex}{\textbf{IoU}} & \raisebox{-0.2ex}{\textbf{GIoU}} \\ 
 \hline
 \hline
$\ell_1$-smooth~\cite{faster_rcnn}  & .360 & .351 && .390 & .379 \\
 \hline
 \hline
 $\calL_{IoU}$ & .368 & .358 && .396 & .385 \\
 Relative improv.\% & 2.22\% & 1.99\% && 1.54\% & 1.58\% \\
 \hline
  $\calL_{GIoU}$ & \textbf{.369} & \textbf{.360} && \textbf{.398} & \textbf{.388} \\
  Relative improv. \% & \textbf{2.50\%} & \textbf{2.56\%} && \textbf{2.05\%} & \textbf{2.37\%} \\
 \hline
 \end{tabular}
 \label{table:faster-rcnn-cocoval}
 %\vspace{-1em}
\end{table}

\begin{table}[!tb]\footnotesize
\centering
   \caption{\footnotesize Comparison between the performance of \textbf{Faster R-CNN}~\cite{faster_rcnn} trained using its own loss ($\ell_1$-smooth) as well as $\calL_{IoU}$ and $\calL_{GIoU}$ losses. The results are reported on the \textbf{test set of MS COCO 2018}.}
 \begin{tabular}{c c c c} 
 \hline 
Loss \big/ Metric & AP & AP75 \\ [0.5ex] 
 \hline
 \hline
 $\ell_1$-smooth~\cite{faster_rcnn}& .364 & .392\\
 \hline
 \hline
  $\calL_{IoU}$ & \textbf{.373} & .403 \\
 Relative improv.\% & \textbf{2.47\%} & 2.81\% \\
 \hline
 $\calL_{GIoU}$ & \textbf{.373} & \textbf{.404} \\
  Relative improv.\% & \textbf{2.47\%} & \textbf{3.06\%} \\
 \hline
 \end{tabular}
  \label{table:faster-rcnn-cocotest}
\end{table}

\begin{table}[!tb]\footnotesize
\centering
   \caption{\footnotesize Comparison between the performance of \textbf{Mask R-CNN}~\cite{he2017mask} trained using its own loss ($\ell_1$-smooth) as well as $\calL_{IoU}$ and $\calL_{GIoU}$ losses. The results are reported on the \textbf{validation set of MS COCO 2018}.}
 \begin{tabular}{c c  c c c c} 
 \hline 
\raisebox{-1.5ex}{Loss \big/ Evaluation} & \multicolumn{2}{c}{\raisebox{-0.5ex}{AP}} &&\multicolumn{2}{c}{\raisebox{-0.5ex}{AP75}}\\ [0.5ex] 
\cline{2-3}\cline{5-6}
 & \raisebox{-0.2ex}{\textbf{IoU}} & \raisebox{-0.2ex}{\textbf{GIoU}} &&\raisebox{-0.2ex}{\textbf{IoU}} & \raisebox{-0.2ex}{\textbf{GIoU}} \\ 
 \hline
 \hline
$\ell_1$-smooth~\cite{he2017mask}  & .366 & .356 && .397 & .385 \\
 \hline
 \hline
 $\calL_{IoU}$ & .374 & .364 && .404 & .393 \\
 Relative improv.\% & 2.19\% & 2.25\% && 1.76\% & 2.08\% \\
 \hline
  $\calL_{GIoU}$ & \textbf{.376} & \textbf{.366} && \textbf{.405} & \textbf{.395} \\
  Relative improv. \% & \textbf{2.73\%} & \textbf{2.81\%} && \textbf{2.02\%} & \textbf{2.60\%} \\
 \hline
 \end{tabular}
 \label{table:mask-rcnn-cocoval}
\end{table}

\begin{table}[!tb]\footnotesize
\centering
   \caption{\footnotesize Comparison between the performance of \textbf{Mask R-CNN}~\cite{he2017mask} trained using its own loss ($\ell_1$-smooth) as well as $\calL_{IoU}$ and $\calL_{GIoU}$ losses. The results are reported on the \textbf{test set of MS COCO 2018}.}
 \begin{tabular}{c c c c} 
 \hline 
Loss \big/ Metric & AP & AP75\\ [0.5ex] 
 \hline
 \hline
 $\ell_1$-smooth~\cite{he2017mask}& .368 & .399 \\
 \hline
 \hline
  $\calL_{IoU}$ & \textbf{.377} & .408 \\
 Relative improv.\% & \textbf{2.45\%} & 2.26\% \\
 \hline
 $\calL_{GIoU}$ & \textbf{.377} & \textbf{.409} \\
  Relative improv.\% & \textbf{2.45\%} & \textbf{2.51\%} \\
 \hline
 \end{tabular}
  \label{table:mask-rcnn-cocotest}
\end{table}

\begin{figure*}[!tbh]
\begin{center}
	\frame{\includegraphics[width=1\linewidth]{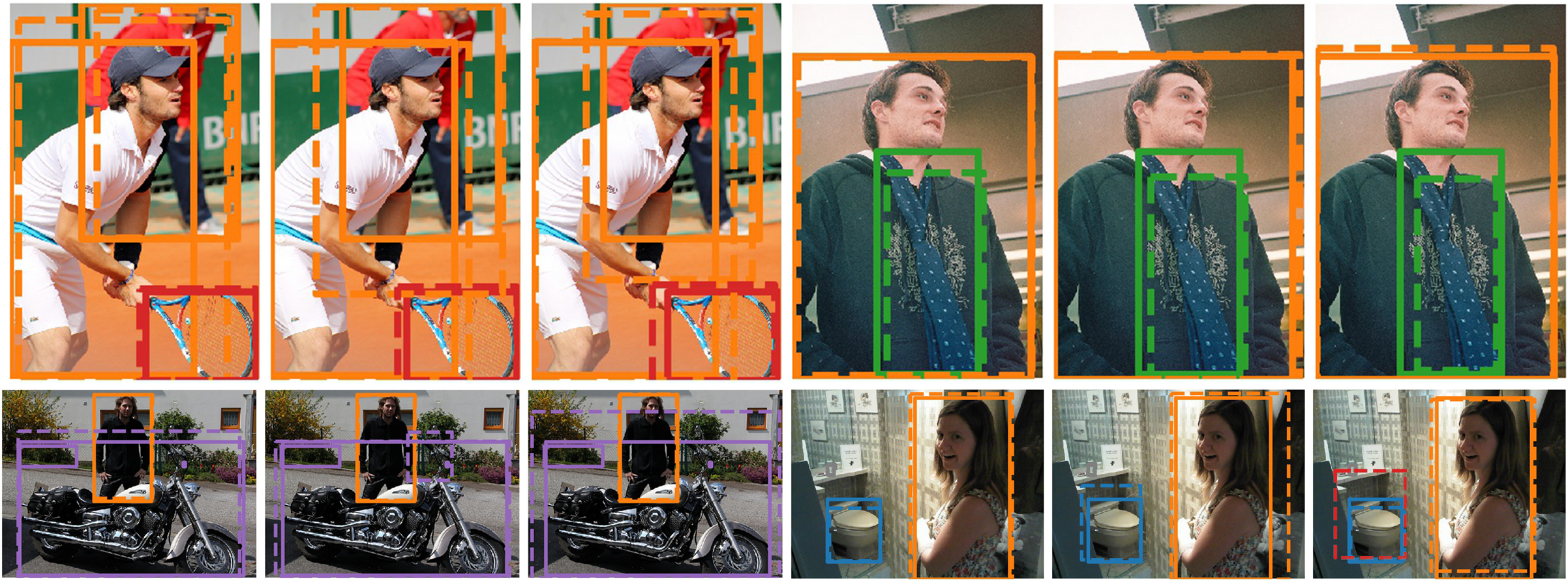}}
\end{center}
\vspace{-1em}
\caption{Example results from COCO validation using YOLO v3~\cite{yolov3} trained using (left to right) $\calL_{GIoU}$, $\calL_{IoU}$, and MSE losses. Ground truth is shown by a solid line and predictions are represented with dashed lines.}
\end{figure*}

\begin{figure*}[!tbh]
\begin{center}
	\frame{\includegraphics[width=1\linewidth]{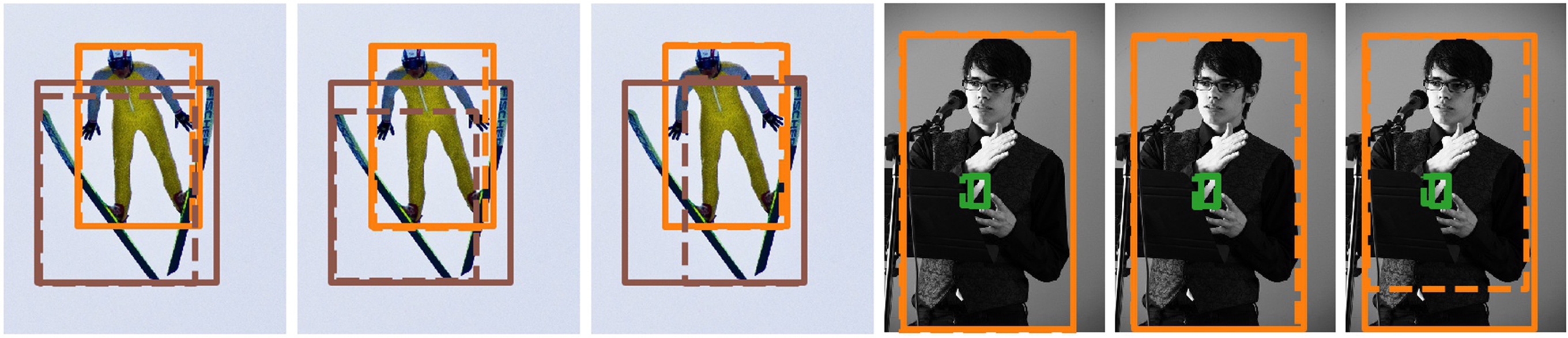}}
\end{center}
\vspace{-1em}
\caption{Two example results from COCO validation using Mask R-CNN~\cite{he2017mask} trained using (left to right) $\calL_{GIoU}$, $\calL_{IoU}$, $\ell_1$-smooth losses. Ground truth is shown by a solid line and predictions are represented with dashed lines.}
\end{figure*}

\section{Conclusion}
In this paper, we introduced a generalization to $IoU$ as a new metric, namely $GIoU$, for comparing any two arbitrary shapes. We showed that this new metric has all of the appealing properties which $IoU$ has while addressing its weakness. Therefore it can be a good alternative in all performance measures in 2D/3D vision tasks relying on the $IoU$ metric. 

We also provided an analytical solution for calculating $GIoU$ between two axis-aligned rectangles. We showed that the derivative of $GIoU$ as a distance can be computed and it can be used as a bounding box regression loss. By incorporating it into the state-of-the art object detection algorithms, we consistently improved their performance on popular object detection benchmarks such as PASCAL VOC and MS COCO using both the commonly used performance measures and also our new accuracy measure, \ie $GIoU$ based average precision. Since the optimal loss for a metric is the metric itself, our $GIoU$ loss can be used as the optimal bounding box regression loss in all applications which require 2D bounding box regression.

In the future, we plan to investigate the feasibility of deriving an analytic solution for $GIoU$ in the case of two rotating rectangular cuboids. This extension and incorporating it as a loss could have great potential to improve the performance of 3D object detection frameworks.
%------------------------------------------------------------------------

{\small
\bibliographystyle{ieee}
\bibliography{egbib}
}

\end{document}